\documentclass[runningheads]{llncs}
\usepackage[T1]{fontenc}
\usepackage{graphicx}
\usepackage{amsmath}
\usepackage{hyperref}

\usepackage[acronym]{glossaries}
\usepackage{multirow}
\usepackage{arydshln}
\usepackage{adjustbox}
\usepackage{booktabs}
\usepackage{todonotes}
\usepackage{xspace}
\usepackage{dsfont}

\newacronym{egl}{EGL}{Explanation-Guided Learning}
\newcommand{\testid}{\ensuremath{\mathrm{Test}_\mathrm{ID}}\xspace}
\newcommand{\testrand}{\ensuremath{\mathrm{Test}_\mathrm{rand}}\xspace} 
\newcommand{\visa}{\ensuremath{\mathrm{VisFIS}^\ast}\xspace}
\newcommand{\last}{\ensuremath{\lambda^\ast}\xspace}
\newcommand{\vit}{ViT\xspace}
\newcommand{\supplink}{\url{https://right-for-wrong-reasons.dfki.de/}}

\begin{document}

\title{Right for the Wrong Reasons?\\
Evaluating Plausibility and Behavioral Change\\in Explanation-Guided Learning%
\thanks{%
Presented at EXPLAINS 2026 (Best Student Paper).
For the latest manuscript version and
supplementary material, visit:}
}

\titlerunning{Right for the Wrong Reasons?}
%
\author{David Dembinsky\inst{1,2} \and
Adriano Lucieri\inst{1} \and
Andreas Dengel\inst{1,2}}
\authorrunning{D. Dembinsky et al.}
%
\institute{German Research Center for Artificial Intelligence (DFKI) GmbH, Germany, 
\\ \email{\{firstname\}.\{lastname\}@dfki.de}\and
RPTU University Kaiserslautern-Landau, Germany
}

\maketitle

\begin{abstract}
\gls{egl} promises models that are not only accurate but also rely on task-relevant evidence.
However, whether improved agreement between model explanations and human-provided rationales also implies a changed decision strategy remains unclear.
We study this tension under controlled synthetic shortcut introduction using a constructed shortcut dataset. 
Inspired by real-world applications, our benchmark covers shortcuts of different kinds, including localized, background-based, and global cues.
This allows us to evaluate whether different \gls{egl} objectives produce behavioral changes
and explanatory changes.
Across these axes, we find that the effects of \gls{egl} are strongly objective-specific.
Rationale-based data augmentation provides the clearest robustness gains when shortcut evidence can be separated from task-relevant foreground information, reaching average accuracy improvements of up to \(+4.8\) percentage points (pp) together with increased foreground reliance.
Attribution-constraining objectives primarily improve explanation plausibility, increasing Importance Mass by up to \(+14\) pp on average, but provide weaker and less consistent behavioral correction, with average accuracy gains of at most \(+1.6\) pp.
Moreover, neither improved robustness nor improved plausibility reliably implies higher perturbation-based fidelity.
Overall, our results show that \gls{egl} can be useful, but its success cannot be inferred from rationale agreement alone and should instead be evaluated along complementary behavioral and explanation-level axes.
We publicly release all experimental results to enable further analysis.\footnote{\supplink}

\end{abstract}

\section{Introduction}
Ever more complex machine learning models are deployed in high-stakes domains such as medicine and finance, where opaque decision processes complicate human oversight and may obscure biases encoded by the model.
In response, the community has developed a broad range of methods for explaining model predictions.
This growing emphasis on transparency is also reflected in regulation such as
the EU AI Act, which establishes transparency requirements for high-risk AI
systems \cite{europeancommission2021aiactproposal}.

While explanations are commonly used as post-hoc diagnostic tools, transparency can also be pursued during model design or training, for example through inherently interpretable architectures.
A complementary strategy is to retain flexible models but guide them during training toward user-specified evidence.
This idea underlies \emph{Explanation-Guided Learning} (\gls{egl}), which augments the classification loss with auxiliary objectives derived from user-provided rationales \cite{Gao2024going,ross2017right}.
In its classical form, such guidance directly constrains model explanations to align with these rationales, aiming to encourage models to base their predictions on task-relevant evidence and follow the aspiration of being ``right for the right reasons.''
However, whether optimizing explanations actually changes the model's decision strategy in this intended way remains an empirical question.
The promise of explanation-guided training has led to a growing body of work in computer vision and natural language processing \cite{Gao2024going,selvaraju2019taking,simpson2019gradmask,viviano2021saliency,ying2022visfis}.

Assessing whether \gls{egl} fulfills its intended goals is further complicated by a gap between the apparent quality of an explanation and the model behavior it is meant to reflect.
\emph{Plausibility}, i.e. agreement between an explanation and human annotations, does not necessarily imply \emph{fidelity}, i.e. causal relevance of the highlighted features to the model's decision \cite{dembinsky2026unifying,hase2020evaluating}\footnote{The terms \textit{faithfulness} and \textit{fidelity} are used interchangeably in this work.}.
A model may therefore be trained to produce plausible explanations while still relying on shortcuts or spurious correlations \cite{hase2020leakage}.
For \gls{egl}, this raises a central question: do objectives that improve rationale agreement also induce the intended behavioral changes, such as greater robustness, reduced shortcut reliance, and more faithful explanations, or do they primarily optimize the appearance of explanations?
This question is especially relevant when comparing classical attribution-constraining objectives with rationale-based data-augmentation objectives, since both use rationales during training but may affect the learned model differently.
Yet prior work has mostly considered isolated shortcut settings and often emphasizes plausibility metrics, leaving the relation between objective type, model behavior, and explanation quality insufficiently understood.

\paragraph{Contributions.}
We conduct a multi-axis evaluation of \gls{egl} under controlled synthetic shortcut bias.
To this end, we construct a benchmark that injects three shortcut types: localized patch cues, class-correlated backgrounds, and global color-cast cues.
Using this benchmark, we compare several \gls{egl} objectives for a convolutional and a transformer-based model, distinguishing attribution-constraining objectives from rationale-based data-augmentation objectives.
We evaluate the resulting models along four complementary axes, grouped into behavioral and explanation-level effects: robustness to shortcut shift (\hyperref[sec:rq1]{RQ1}) and reliance on foreground versus shortcut evidence (\hyperref[sec:rq2]{RQ2}) as behavioral axes, and explanation plausibility with respect to object masks (\hyperref[sec:rq3]{RQ3}) and explanation fidelity to the model's decision process (\hyperref[sec:rq4]{RQ4}) as explanation-level axes.
Our results show that these desiderata can diverge substantially.
The Attribution-constraining objectives yield the clearest plausibility gains, whereas rationale-based data-augmentation objectives are more effective for improving robustness and reducing shortcut reliance in separable shortcut settings.
These findings show that evaluating \gls{egl} primarily through explanation plausibility is insufficient: improved rationale agreement does not show that the model relies on the intended features, and objectives that directly intervene on the training data may be more effective for changing model behavior than attribution alignment.

\section{Related Work}
\label{sec:related_work}

\gls{egl} provides a broad family of approaches that aim to shape model behavior by incorporating explanation-related feedback into training; for a comprehensive overview of objectives, feedback types, and application domains, we refer to \textit{Gao et al.} \cite{Gao2024going}.
In the classical attribution-based formulation, \gls{egl} augments standard task supervision with an additional explanation-supervision term:
\begin{equation}
    \min \underbrace{\mathcal{L}_\mathrm{Pred}\Big(f(X),Y\Big)}_\text{task supervision} 
    \,+\, 
    \underbrace{\lambda\mathcal{L}_\mathrm{Exp}\Big(\xi(f,X,Y),\tilde{E}\Big)}_\text{explanation supervision} 
    \qquad,
\end{equation}
where an explanation function $\xi$ produces an explanation for the model $f$, which is compared against a supposed ground-truth rationale $\tilde{E}$ \cite{Gao2024going}.\footnote{In their survey, \textit{Gao et al.} additionally include an \textit{explanation regularization} term \cite{Gao2024going}, which we omit for simplicity, as it is not central to the present work.}
In computer vision and vision-language reasoning, such ground-truth rationales $\tilde{E}$ are commonly represented as bounding boxes or segmentation masks \cite{selvaraju2019taking,viviano2021saliency,ying2022visfis}.
While intuitive, these targets may be noisy, incomplete, or overly coarse, and even accurate spatial annotations need not coincide with the evidence actually used by a neural network.
This raises the concern that optimizing explanations toward human-interpretable localization targets may improve the appearance of explanations without necessarily improving the model's decision process \cite{viviano2021saliency}.
A more domain-specific alternative is the use of causally or physically grounded continuous explanation targets, as in \textit{Talies et al.}, who derive target explanations from the stress field around a crack tip \cite{talies2026trustworthy}.

The explanation function $\xi$ is usually chosen to be differentiable, since the explanation loss must be optimized during training \cite{Gao2024going}.
Prominent choices include input gradients or Saliency \cite{simonyan2014deep}, as used by \textit{Ross et al.} and \textit{Simpson et al.} \cite{ross2017right,simpson2019gradmask}, and Grad-CAM \cite{selvaraju2017grad}, as used by \textit{Selvaraju et al.} and \textit{Talies et al.} \cite{selvaraju2019taking,talies2026trustworthy}.
These classical explanation-guidance objectives directly constrain the explanation signal, for example by aligning it with the annotated rationale or by penalizing attribution outside the rationale.
VisFIS is particularly relevant for the present work because it combines such attribution-constraining terms with rationale-based data-augmentation terms \cite{ying2022visfis}.
Its attribution-based terms align explanations with ground-truth feature-importance annotations and penalize saliency outside the annotated rationale, whereas its sufficiency and uncertainty objectives use the rationale to construct augmented training inputs by preserving or removing rationale regions.
Thus, VisFIS provides a natural objective family for separating explanation-alignment effects from rationale-based data-augmentation effects: both use ``ground-truth'' rationales during training, but only the former directly optimizes explanations, while the latter changes the available input evidence.

Empirically, \gls{egl} is often reported to improve predictive performance, but the effect is neither uniform nor independent of the chosen explanation target or objective type.
In standard supervised settings, explanation guidance can lead to moderate performance improvements, for instance in medical imaging or crack-tip segmentation, but can also reduce performance in individual settings \cite{simpson2019gradmask,talies2026trustworthy}.
A central motivation is its potential to mitigate shortcut learning and improve generalization under distribution shift, which has been studied in controlled settings with known irrelevant features \cite{ross2017right}, vision-language benchmarks with shifted answer distributions \cite{selvaraju2019taking,ying2022visfis}, and image-classification settings with spatial confounders \cite{viviano2021saliency}.
Overall, these works indicate that explanation-related supervision can reduce reliance on certain shortcuts, but the evidence remains limited to a relatively small set of shortcut types.
Moreover, \textit{Viviano et al.} show that attribution priors may improve generalization under covariate shift, but can also hurt performance when no shift is present, and that improved attribution localization does not necessarily imply improved out-of-distribution performance \cite{viviano2021saliency}.
Thus, it remains unclear whether the observed benefits transfer to broader shortcut scenarios, and whether they are driven by explanation alignment itself or by objectives that intervene on the training data.

The effect of \gls{egl} on the resulting post-hoc explanations is evaluated less consistently than its effect on predictive performance.
Most works primarily assess plausibility, i.e. whether explanations overlap with human- or domain-provided target regions \cite{ross2017right,selvaraju2019taking,simpson2019gradmask,ying2022visfis}.
Across these studies, \gls{egl} generally improves agreement with the annotated rationale, but this alone does not establish that the explanation faithfully represents the model's decision process.
Only a smaller subset of works evaluates explanation fidelity explicitly, for example through comparisons with occlusion-based importance \cite{selvaraju2019taking}, sufficiency and comprehensiveness \cite{ying2022visfis}, or deletion- and insertion-based metrics \cite{talies2026trustworthy}.
These studies suggest that plausibility, fidelity, and predictive performance are related but distinct: an increase in explanation plausibility can only be expected to improve predictive performance if the explanation is also faithful to the model's behavior \cite{ying2022visfis}. 

Taken together, prior work motivates \gls{egl} as a promising strategy for improving generalization and steering model behavior, but two limitations remain.
First, shortcut mitigation has mostly been investigated in isolated benchmark settings rather than through a systematic comparison of different shortcut paradigms.
Second, it remains insufficiently clear whether objectives that improve explanation alignment also change the evidence used by the model, or whether they can merely produce explanations that better resemble the provided rationale.
These limitations motivate a controlled evaluation of classical attribution-constraining objectives and rationale-based data-augmentation objectives across shortcut-shift robustness, shortcut reliance, explanation plausibility, and explanation fidelity.

%
%
%
%
%
%

\section{Experiments}
\label{sec:experiments}
This section describes the experimental setup used to evaluate \gls{egl} under controlled shortcut bias.
To keep the main text readable, we report representative results and provide the full experimental data through an interactive website at \supplink.

\subsubsection*{Setup}
\paragraph{Dataset.}
To evaluate \gls{egl} under controlled shortcut bias, we construct a novel, customized-FMNIST (cFMNIST), a synthetic benchmark inspired by previous work \cite{bykov20222noisegrad} and based on Fashion-MNIST foreground objects \cite{xiao2017fashion} placed on CIFAR-10 backgrounds \cite{krizhevsky2009learning}.
The task is to classify the foreground object, while shortcut cues are introduced independently.

Real-world vision models may exploit shortcuts that are predictive in the training data but unrelated to the intended task, including artifacts observed in medical imaging.
We therefore consider three shortcut variants: \textit{cFMNIST-patch}, where a colored square in the upper-left corner serves as a localized shortcut, reflecting marker-like or border artifacts reported in skin-lesion analysis \cite{bissoto2020debiasing,sies2021dark}; \textit{cFMNIST-background}, where the CIFAR-10 background class is correlated with the foreground label, reflecting background and context biases in vision datasets \cite{ribeiro2016should,xiao2021noise}; and \textit{cFMNIST-color}, where a class-correlated color cast is applied to the full image, abstracting acquisition- or site-specific global shifts \cite{howard2021impact,soylu2023calibrating}.
We additionally include \textit{cFMNIST-clean}, which uses the same foreground-background construction but no class-correlated shortcut.

For each shortcut variant, we use several different correlation strengths $p\in\{0.5,0.8,0.95\}$, while $p=0.1$ corresponds to random shortcut assignment in the ten-class task and is used for shortcut-independent evaluation.
The reference rationale $\tilde{E}$ is the foreground object mask.
Dataset construction details, split sizes, and diagnostic variants are given in \autoref{app:dataset}.

\paragraph{Models.}
We evaluate \gls{egl} on two complementary architectures: a ResNet18 \cite{he2016deep} trained from scratch and an ImageNet-pretrained ViT-B/16 (\vit) \cite{dosovitskiy2020image}.
ResNet18 serves as a widely used convolutional baseline for vision experiments, while the Vision Transformer represents a more recent architecture family whose attention-based design has become influential across vision and language tasks.
This allows us to test whether the observed effects of \gls{egl} are specific to a standard convolutional model or also appear in a contemporary transformer-based architecture.

\paragraph{EGL objectives.}
We follow the VisFIS objective family \cite{ying2022visfis}, which combines rationale-based data augmentation with classical explanation guidance and relates to earlier attribution-guided training approaches \cite{simpson2019gradmask,talies2026trustworthy,viviano2021saliency}.
We evaluate four objectives individually, grouped into two families.
Rationale-based data-augmentations modify the input seen during training: \emph{Sufficiency} trains the model on a foreground-only view, while \emph{Uncertainty} discourages confident predictions when the foreground is removed.
Attribution-constraining objectives instead act on the explanation itself: \emph{Invariance} penalizes attribution outside the foreground rationale, while \emph{Alignment} directly aligns the attribution map with the foreground mask.
We additionally evaluate a combined \visa objective using the individually selected coefficients for the four terms, without an additional joint hyperparameter search.

Unless stated otherwise, attribution-constraining objectives are optimized using Vanilla Gradient as the training-time explanation signal.
We follow Ying et al.~\cite{ying2022visfis}, who report Vanilla Gradient to perform best among their considered guidance signals, while noting that the differences between guidance signals are modest.
For each objective, we train a grid of $\lambda$ values and report medians across random seeds; full objective definitions, selection criteria, and training details are provided in \autoref{app:training_details} and \autoref{app:explanation_guidance}.

%
%
%
%
%
%

\begin{figure}[t]
\centering
\includegraphics[width=1\linewidth]{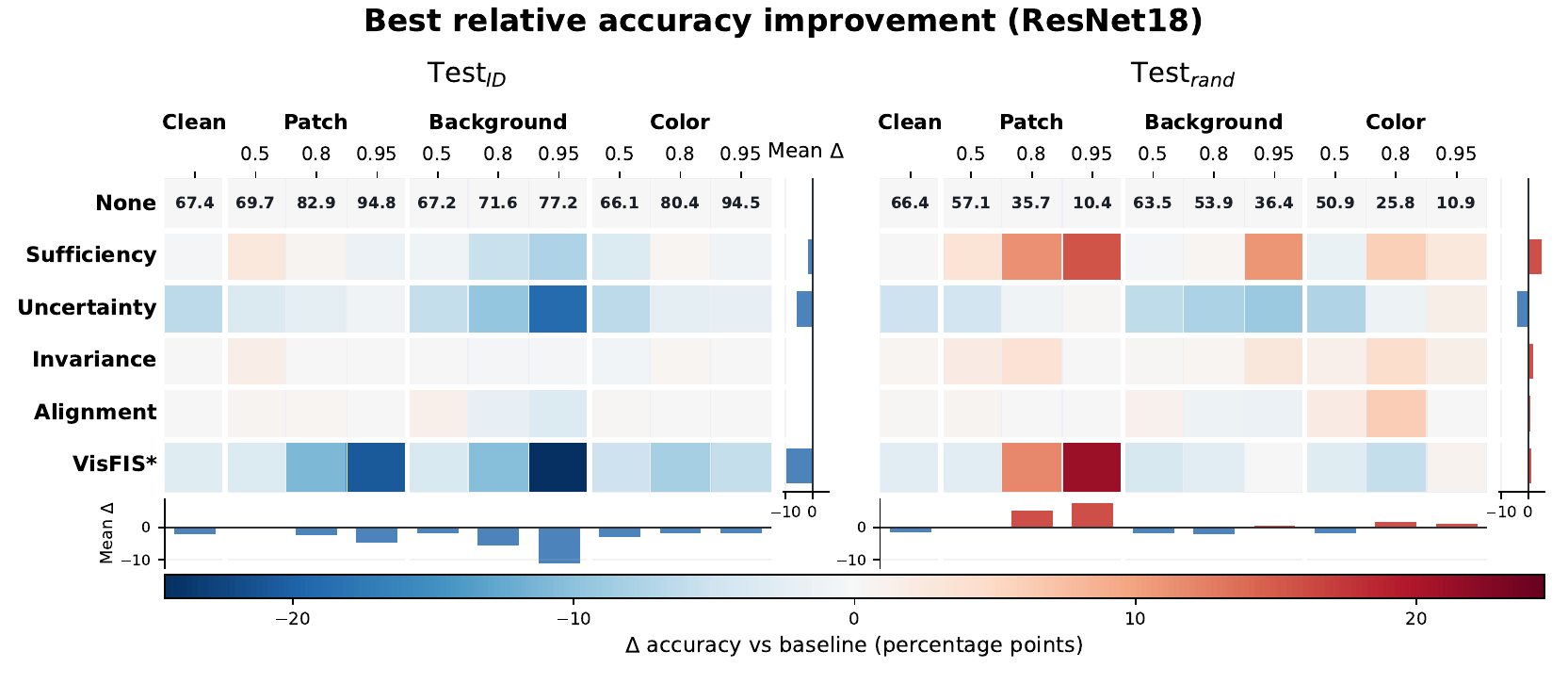}
\caption{
Relative accuracy change of ResNet18 under \gls{egl} compared with the unguided baseline.
For each objective and dataset setting, \last is selected by highest \testrand accuracy; both the resulting \testrand change and the corresponding \testid change are reported for the same \last.
Columns group shortcut types and correlations $p$, rows show \gls{egl} objectives, and colors indicate accuracy changes in percentage points.
Baseline accuracies are shown above each column.
}
\label{fig:rq1_accs_resnet}
\end{figure}

\subsection{RQ1: Which EGL objectives improve generalization under shortcut shift?}
\label{sec:rq1}

RQ1 evaluates which \gls{egl} objectives improve generalization when shortcut correlations present during training are removed at test time.
We compare two test splits: \testid follows the same shortcut correlation as the training data, whereas \testrand uses random shortcut assignment with $p=0.1$.
For the \textit{Clean} dataset, which contains no shortcut correlation during training, both test splits are interchangeable.

\autoref{fig:rq1_accs_resnet} summarizes the ResNet18 results as accuracy changes relative to the unguided baseline.
Columns correspond to shortcut types and shortcut strengths, while rows correspond to \gls{egl} objectives.
For each objective and dataset setting, we select \last by highest \testrand accuracy and report both the resulting \testrand change and the corresponding \testid change for the same model.
We focus on ResNet18 in the main text and provide the corresponding \vit results on the supplementary website.
We summarize the main findings below.

\paragraph{The benchmark induces the intended shortcut reliance.}
For the unguided baseline, stronger shortcut correlations increase \testid accuracy but reduce \testrand accuracy.
In \autoref{fig:rq1_accs_resnet}, this is visible from the baseline accuracies above the columns: as $p$ increases, \testid accuracy rises for the shortcut-correlated datasets, while \testrand accuracy drops, most strongly for Patch and Color.
This confirms that the synthetic shortcuts are not merely present in the data, but are actually used by the baseline model.
The resulting \testid--\testrand gap provides the failure mode that \gls{egl} is expected to mitigate.

\paragraph{\gls{egl} is not a general-purpose robustness booster.}
The \testrand block shows localized positive regions rather than uniform improvements across objectives and shortcut types.
In the clean setting, \gls{egl} produces no noteworthy improvement, and for weak shortcut correlations the gains are similarly small.
This is expected: when the baseline has little shortcut reliance to correct, guidance provides limited benefit.
The clearest gains appear in high-correlation settings, where shortcut reliance is strongest and the baseline generalization gap is largest.
Thus, \gls{egl} improves robustness mainly when the model has learned a shortcut that the training objective can effectively discourage.

\paragraph{Sufficiency is strongest for separable shortcuts.}
Among the individual objectives, Sufficiency yields the clearest \testrand improvements for patch shortcuts and high-correlation background shortcuts.
This matches the structure of the objective: masking to the foreground preserves the task-relevant object while removing localized patches or separable background evidence.
For color shortcuts, however, the gains are weaker and less systematic, because the shortcut is distributed across the whole image and cannot be cleanly removed by foreground masking.
In this setting, the attribution-constraining objectives Invariance and Alignment become comparatively more competitive, although their improvements remain smaller and less consistent than the strongest Sufficiency gains.
Uncertainty is the least reliable objective and rarely improves robustness.

\paragraph{Robustness gains often trade off against \testid accuracy.}
Several improvements on \testrand coincide with neutral or negative changes on \testid, especially at stronger shortcut correlations.
This should not be interpreted as a simple failure: \testid still rewards shortcut usage, so a model that relies less on the shortcut can lose in-distribution accuracy while becoming more robust under shortcut shift.
This mirrors prior observations that attribution priors can improve generalization under distribution shift while hurting performance when no shift is present \cite{viviano2021saliency}.
The \visa objective is also not automatically additive.
Although the original paper reports the strongest OOD gains for the combined objective \cite{ying2022visfis}, this pattern does not transfer uniformly to our controlled shortcut setting.
Here, the \visa improves some patch settings, but does not consistently preserve the strongest individual effects and often incurs comparatively large \testid reductions.
This suggests interference between objectives rather than a simple accumulation of benefits.

\paragraph{Answer to RQ1.}
The supplementary \vit results support the same qualitative conclusion, with a generally stronger baseline and less room for improvement.
Thus, RQ1 can be answered positively only in a qualified sense: \gls{egl} can improve shortcut-shift generalization, but the effect is objective-dependent.
The clearest gains come from the rationale-based data-augmentation objective Sufficiency in settings where the shortcut can be suppressed without removing foreground evidence, whereas attribution-constraining objectives provide weaker and less consistent robustness gains.

%
%
%
%
%

\subsection{RQ2: Which EGL objectives reduce shortcut reliance?}
\label{sec:rq2}

While \hyperref[sec:rq1]{RQ1} evaluates which \gls{egl} objectives improve generalization under shortcut shift, RQ2 asks whether these gains are accompanied by a corresponding shift in the evidence used by the model.
We therefore evaluate trained models on masked diagnostic inputs that isolate either the task-relevant foreground or the shortcut cue.
Because foreground and shortcut evidence are controlled by construction, this diagnostic provides a direct behavioral measure of shortcut reliance; details are given in \autoref{app:data_fgbg}.

A successful \gls{egl} objective should increase reliance on foreground evidence relative to shortcut evidence.
We quantify this using the \textit{Foreground-Preference}
\begin{align}
    \Phi_\lambda 
    &= A_{\mathrm{FG},\lambda} - A_{\mathrm{SC},\lambda},
\end{align}
where $A_{\mathrm{FG},\lambda}$ denotes accuracy on foreground-only inputs and $A_{\mathrm{SC},\lambda}$ denotes accuracy on shortcut-only inputs.
Positive values indicate foreground-dominated performance, whereas negative values indicate shortcut-dominated performance.
For consistency with \hyperref[sec:rq1]{RQ1}, we evaluate the \last selected by highest \testrand accuracy; reliance-oriented selections are reported on the supplementary website as an optimistic estimate of attainable foreground-preference shifts.

\begin{figure}[t]
    \centering
    \includegraphics[width=1\linewidth]{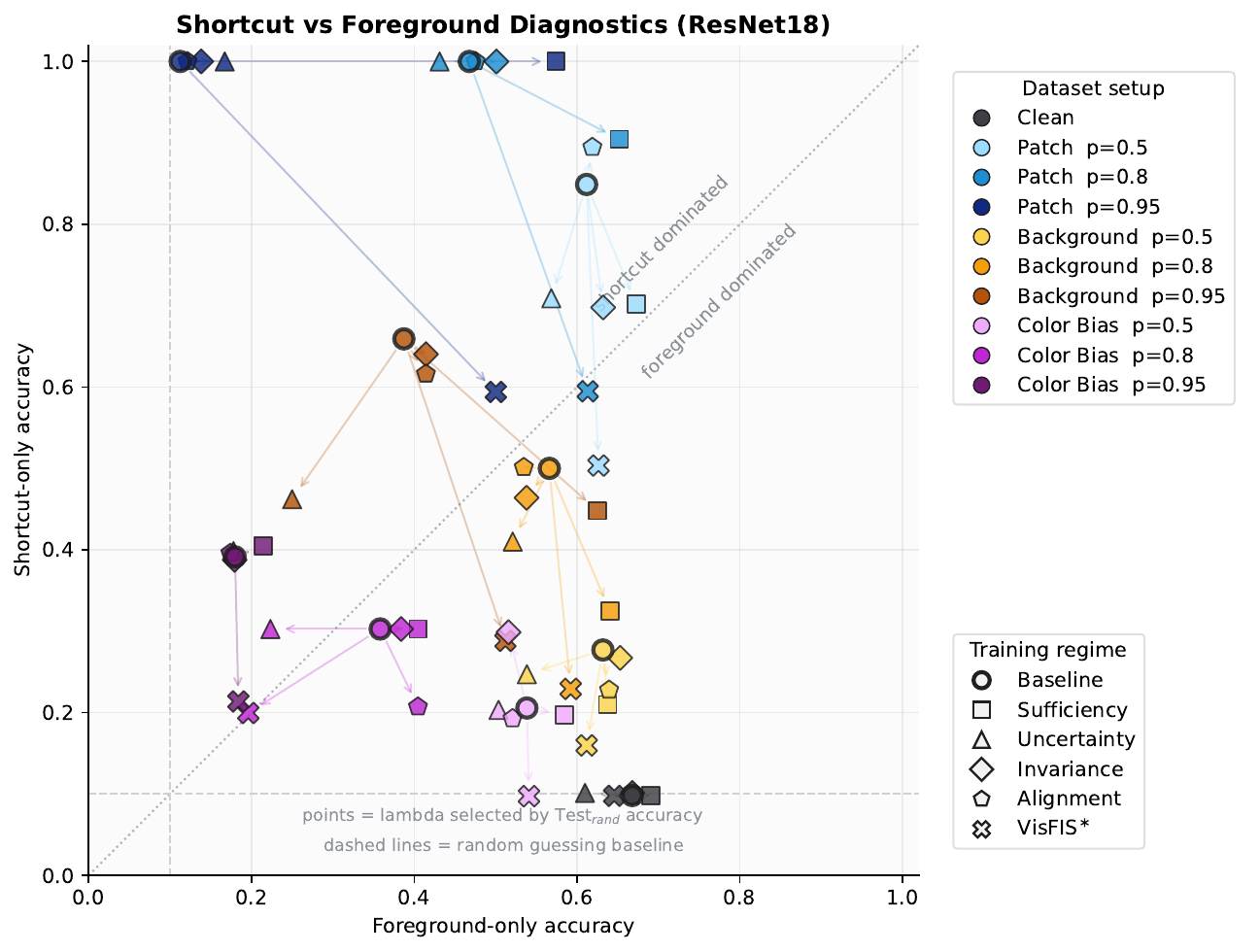}
    \caption{
    Foreground-only versus shortcut-only diagnostic accuracy for ResNet18.
    Each arrow connects the unguided baseline to the \gls{egl} configuration selected by highest \testrand accuracy in \hyperref[sec:rq1]{RQ1}.
    Movement to the right indicates higher foreground-only accuracy; movement downward indicates lower shortcut-only accuracy; movement toward the lower-right therefore indicates stronger foreground preference.
    Dashed lines indicate random-guessing accuracy.
    }
    \label{fig:rq2_scatter_res}
\end{figure}

\autoref{fig:rq2_scatter_res} reports the ResNet18 diagnostics for the robustness-selected models from \hyperref[sec:rq1]{RQ1}.
Each arrow connects the unguided baseline to the corresponding \gls{egl} configuration.
Movement to the right indicates higher foreground-only accuracy, movement downward indicates lower shortcut-only accuracy, and movement toward the lower-right region therefore indicates stronger foreground preference.
Because $\Phi_\lambda$ can increase through higher foreground accuracy, lower shortcut accuracy, or both, we interpret it with the raw diagnostic accuracies shown in the figure.
We focus on ResNet18 in the main text and provide the corresponding \vit results and reliance-oriented $\lambda$ selections on the supplementary website.
We summarize the main findings below.

\paragraph{The diagnostic confirms distinct baseline reliance regimes.}
The ResNet18 baseline shows clear differences in evidence reliance across shortcut types.
In the clean setting, performance is strongly foreground-dominated, as expected, because no shortcut cue is available.
Patch shortcuts, in contrast, are shortcut-dominated across all correlation strengths, with shortcut-only performance becoming especially strong at higher correlations.
For background and color shortcuts, reliance depends more strongly on the shortcut correlation: lower correlations are mostly foreground-oriented, intermediate correlations lie closer to the diagonal, and high-correlation settings become increasingly shortcut-dominated.
This confirms that the diagnostic captures the reliance patterns underlying the shortcut-shift failures observed in \hyperref[sec:rq1]{RQ1}.

\paragraph{Reliance shifts depend on shortcut separability.}
The clearest shifts toward foreground evidence occur for patch shortcuts and high-correlation background shortcuts.
In these settings, shortcut and foreground evidence are spatially separable, so objectives that suppress shortcut evidence can do so without destroying the task-relevant object.
For color shortcuts, the shifts are weaker and less reliable because the shortcut is distributed across the full image and is less cleanly separable from foreground information.
Thus, the diagnostic suggests that reducing shortcut reliance is easiest when the shortcut can be isolated from the foreground evidence.

\paragraph{Sufficiency gives the clearest individual reliance shift.}
Among the individual objectives, the rationale-based data-augmentation objective Sufficiency produces the most reliable movement toward foreground-dominated performance.
This is most visible for patch shortcuts and high-correlation background shortcuts, where foreground-only accuracy increases and shortcut-only accuracy becomes less dominant.
These are also the settings in which Sufficiency yields some of the clearest robustness gains in \hyperref[sec:rq1]{RQ1}, suggesting that improved \testrand performance is often accompanied by a behavioral shift away from shortcut-only evidence.
For color shortcuts, the effect is weaker, consistent with the fact that foreground masking cannot cleanly remove the globally distributed shortcut.

\paragraph{Other objectives produce less systematic reliance changes.}
Uncertainty, Invariance, and Alignment show weaker and less consistent changes under robustness-oriented selection.
Uncertainty can reduce shortcut-only accuracy in some settings, but this often coincides with reduced foreground-only accuracy, making the change difficult to interpret as improved task-relevant reliance.
Invariance and Alignment can improve foreground preference in individual settings, but their effects are generally smaller and less systematic than those of Sufficiency.
The combined \visa objective can yield strong reliance shifts for separable patch and background shortcuts, often by substantially reducing shortcut-only performance, but these shifts are less reliable for color shortcuts and do not always translate into the strongest \testrand gains.

\paragraph{Answer to RQ2.}
Overall, RQ2 supports the interpretation that robustness-selected \gls{egl} models can reduce shortcut reliance, but only under suitable shortcut structures.
The link between robustness gains and reliance shifts is clearest for separable patch and background shortcuts, where Sufficiency and sometimes \visa suppress shortcut evidence while preserving foreground information.
For globally entangled color shortcuts, the correspondence is weaker.
The supplementary \vit results and reliance-oriented selections support the same qualitative conclusion.

%
%
%
%

\subsection{RQ3: Which EGL objectives improve explanation plausibility?}
\label{sec:rq3}

\hyperref[sec:rq1]{RQ1} and \hyperref[sec:rq2]{RQ2} showed that robustness and reliance shifts are strongest for objectives that can suppress shortcut evidence during training.
RQ3 asks whether these same robustness-selected models also produce more plausible post-hoc explanations, or whether plausibility is primarily improved by objectives that directly constrain the explanation signal.
For consistency with \hyperref[sec:rq1]{RQ1}, we evaluate the \last selected by highest \testrand accuracy; plausibility-oriented selections are reported on the supplementary website as an optimistic estimate of attainable plausibility gains.

We focus on \textit{plausibility}, i.e. the alignment between an explanation and a human-expected reference rationale \cite{dembinsky2026unifying}.
In our setting, the foreground mask serves as this reference rationale.
This question is distinct from both robustness and reliance: a model can become more robust without producing more foreground-aligned explanations, and a plausible explanation need not faithfully reflect the model's decision process.
The latter is evaluated separately in \hyperref[sec:rq4]{RQ4}.

\begin{figure}[t]
    \centering
    \includegraphics[width=1\linewidth]{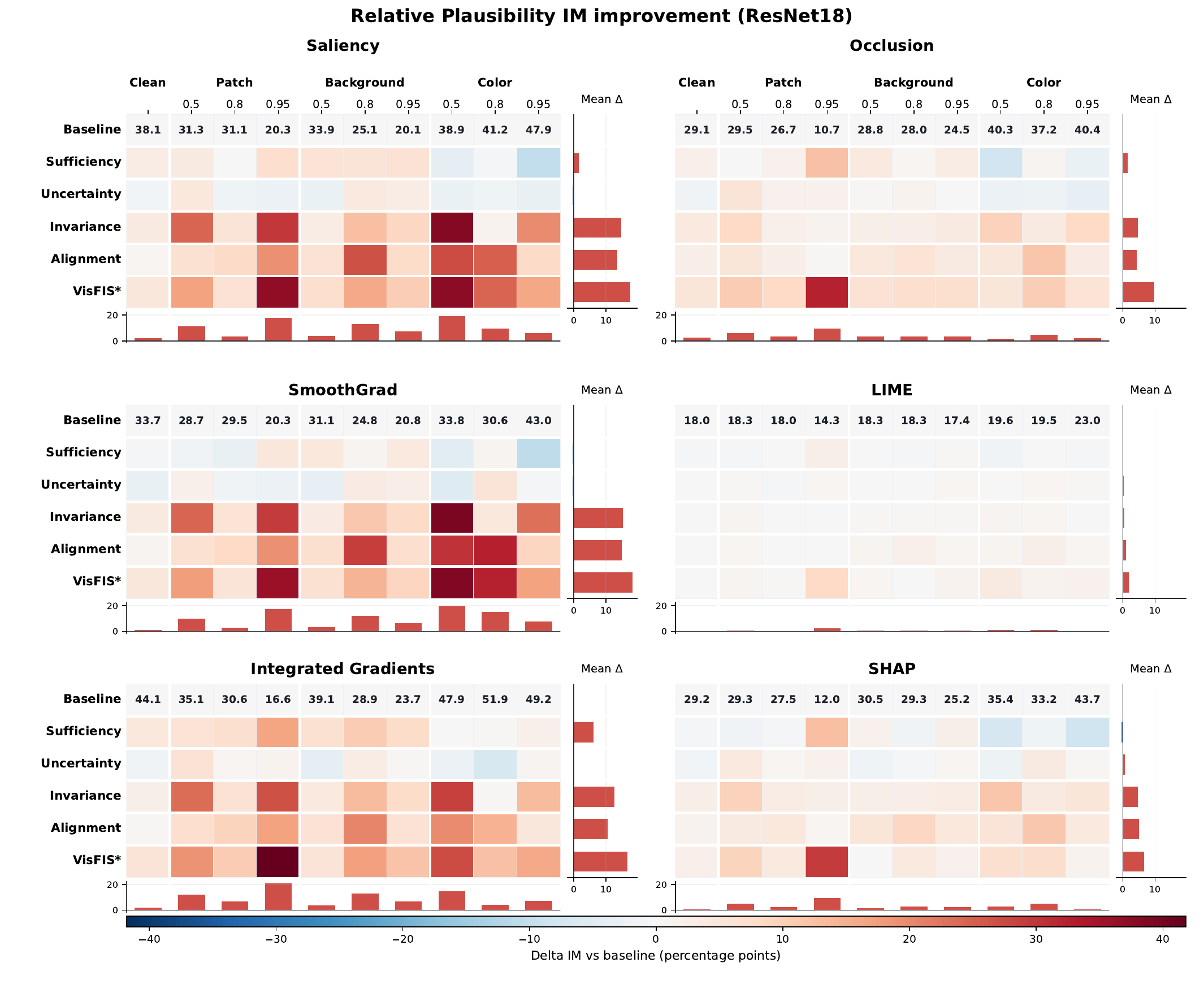}
    \caption{
    Relative change in explanation plausibility for ResNet18, measured by Importance Mass against the foreground rationale.
    For each objective and dataset setting, \last is selected by highest \testrand accuracy as in \hyperref[sec:rq1]{RQ1}.
    Columns group shortcut types and correlations $p$, rows show \gls{egl} objectives, and panels correspond to post-hoc explanation methods.
    Colors indicate changes in percentage points relative to the unguided baseline; baseline Importance Mass values are shown above each column.
    }
    \label{fig:rq3_im_resnet}
\end{figure}

We evaluate six post-hoc explanation methods to test whether explanation guidance affects only the optimized Saliency signal or transfers across explanation families: Saliency \cite{simonyan2014deep}, SmoothGrad \cite{smilkov2017smoothgrad}, Integrated Gradients \cite{sundararajan2017axiomatic}, Occlusion \cite{zeiler2014visualizing}, LIME \cite{ribeiro2016should}, and SHAP \cite{lundberg2017unified}.
Plausibility is measured using Importance Mass (IM), AUROC, and IoU AUC, which respectively capture attribution mass inside the foreground, foreground-background ranking, and spatial overlap across thresholds.
We use Importance Mass as the main metric because it directly measures foreground concentration; details on explanation methods and plausibility metrics are provided in \autoref{app:explanation_methods} and \autoref{app:plausibility_metrics}.
\autoref{fig:rq3_im_resnet} summarizes the ResNet18 plausibility results for Importance Mass under the robustness-selected \last.
We focus on Importance Mass and ResNet18 in the main text and provide the corresponding \vit results, plausibility-oriented $\lambda$ selections, and additional AUROC and IoU AUC analyses on the supplementary website.
We summarize the main findings below.

\paragraph{Plausibility gains transfer unevenly across explanation families.}
Across the six post-hoc explanation methods, the main distinction is between gradient-based explanations and perturbation-based or surrogate explanations.
Saliency, SmoothGrad, and Integrated Gradients show similar qualitative patterns, whereas Occlusion, LIME, and SHAP improve more weakly and less consistently.
This is expected because \gls{egl} uses Saliency as the training-time explanation signal: supervision transfers most reliably to closely related gradient-family explanations and less reliably to explanation methods based on perturbations or local surrogate models.

\paragraph{Attribution-constraining objectives yield the clearest plausibility gains.}
Among the \gls{egl} objectives, Invariance and Alignment produce the strongest and most consistent improvements in Importance Mass.
This matches their objective structure, as both directly constrain the training-time attribution signal to concentrate on the foreground rationale.
In contrast, Sufficiency, which was central for robustness and reliance in \hyperref[sec:rq1]{RQ1} and \hyperref[sec:rq2]{RQ2}, yields more modest plausibility gains, while Uncertainty does not produce a consistent plausibility improvement.
Thus, the objectives that most improve model behavior are not necessarily the objectives that most improve explanation plausibility.

\paragraph{Plausibility gains are driven more by objective type than shortcut type.}
Unlike the robustness and reliance results, plausibility improvements do not follow a simple pattern across shortcut types or shortcut correlations.
This partly reflects the selection strategy: \(\lambda^\ast\) is selected by \testrand accuracy rather than by plausibility, so the heatmap shows the plausibility of robustness-selected models rather than the maximum attainable plausibility for each setting.
Plausibility-oriented selections produce more consistent improvements, but preserve the same qualitative conclusion that Invariance and Alignment are the strongest plausibility-oriented objectives.
The additional AUROC and IoU AUC results also support this interpretation.
Importance Mass and AUROC show the clearest improvements for attribution concentration and foreground-background ranking, whereas IoU AUC is more conservative, suggesting that \gls{egl} more reliably increases attribution mass inside the foreground than it produces uniformly better object-shaped attribution maps.

\paragraph{Answer to RQ3.}
Overall, RQ3 shows that robustness-selected \gls{egl} models do not automatically produce more plausible explanations.
The strongest plausibility gains come from attribution-constraining objectives, especially Invariance and Alignment, whereas the strongest robustness and reliance gains in \hyperref[sec:rq1]{RQ1} and \hyperref[sec:rq2]{RQ2} are driven mainly by Sufficiency in separable shortcut settings.
This extends prior observations that improved attribution localization need not imply improved out-of-distribution performance \cite{viviano2021saliency} and that plausibility should be interpreted through the lens of faithfulness \cite{ying2022visfis}.
In our controlled shortcut setting, improved model behavior does not imply more plausible post-hoc explanations, and improved plausibility does not imply reduced shortcut reliance.
The supplementary \vit results, additional metrics, and plausibility-oriented selections support the same qualitative conclusion.
%
%
%

\subsection{RQ4: EGL objectives improve explanation fidelity?}
\label{sec:rq4}

\hyperref[sec:rq3]{RQ3} shows that attribution-constraining objectives can make explanations more plausible, while \hyperref[sec:rq1]{RQ1} and \hyperref[sec:rq2]{RQ2} show that Sufficiency can improve robustness and foreground reliance in separable shortcut settings.
RQ4 asks whether either effect is accompanied by more faithful post-hoc explanations.
This question is central to the interpretation of \gls{egl}: VisFIS argues that explanation plausibility and predictive performance are meaningfully connected only when explanations are also faithful to the model's behavior \cite{ying2022visfis}.
We therefore test whether models selected for shortcut-shift robustness also yield explanations that better reflect the features relevant to their predictions.

We use the same six post-hoc explanation methods as in \hyperref[sec:rq3]{RQ3}: Saliency, SmoothGrad, Integrated Gradients, Occlusion, LIME, and SHAP.
For feature-attribution explanations, fidelity means that attribution scores correctly identify the features relevant to the model's prediction \cite{dembinsky2026unifying}.
We evaluate fidelity using Decaying Degradation Score (DDS) \cite{vsimic2022perturbation}, a guided perturbation fidelity metric in the terminology of the VXAI framework \cite{dembinsky2026unifying}.
DDS tests whether the ranking induced by an explanation agrees with the model's behavior under feature perturbation: features assigned high relevance should affect the model output earlier and more strongly than features assigned low relevance.
Fidelity is evaluated with respect to the predicted class, so that the metric targets the decision actually explained by the attribution map.
Formal definitions and implementation details are provided in \autoref{app:fidelity_metrics}.

\begin{figure}[t]
    \centering
    \includegraphics[width=1\linewidth]{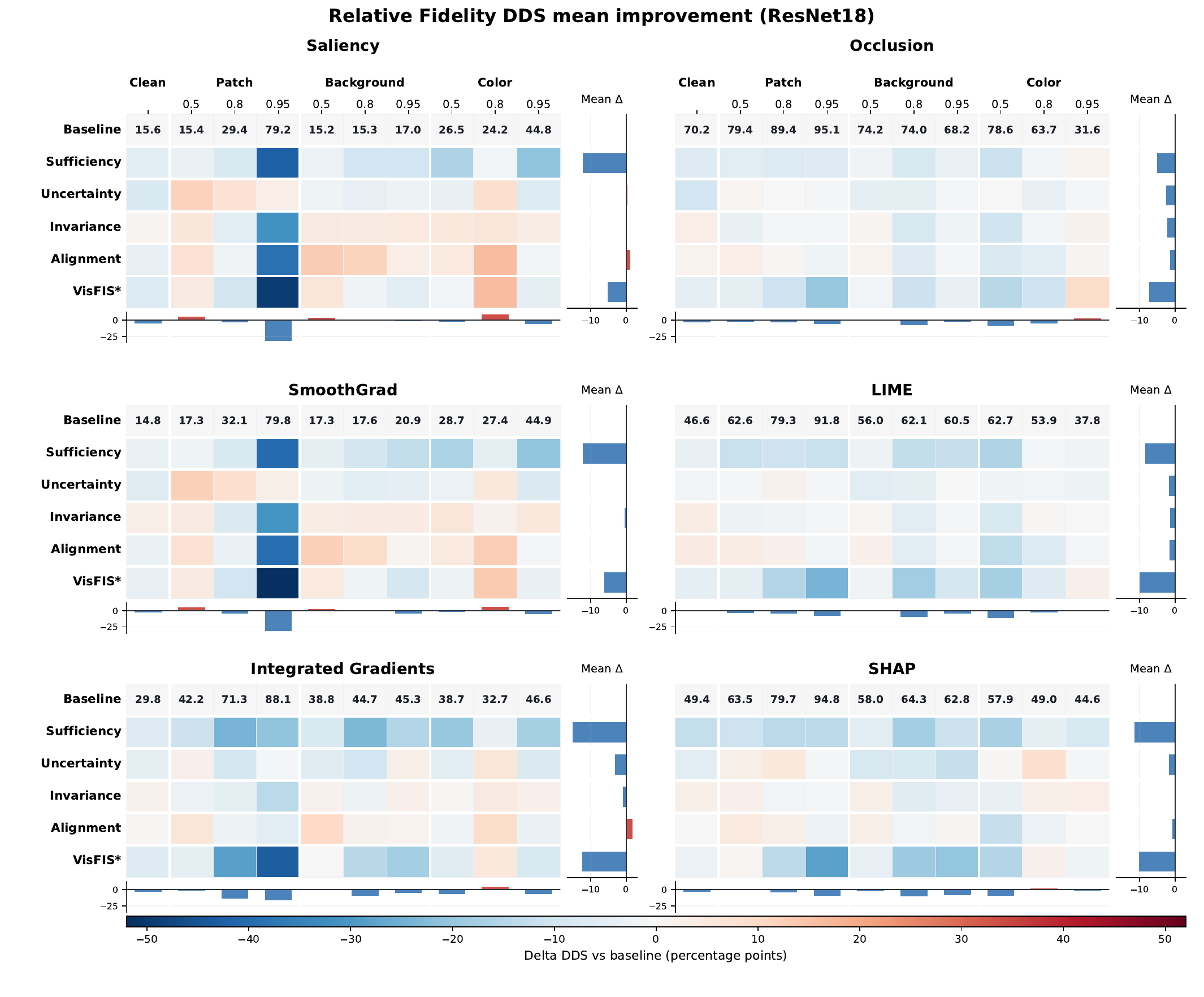}
    \caption{
    Relative change in explanation fidelity for ResNet18, measured by DDS with mean replacement.
    For each objective and dataset setting, $\lambda^\ast$ is selected by highest \testrand accuracy as in \hyperref[sec:rq1]{RQ1}.
    Columns group shortcut types and correlations $p$, rows show \gls{egl} objectives, and panels correspond to post-hoc explanation methods.
    Colors indicate changes relative to the unguided baseline; baseline DDS values are shown above each column.
    }
    \label{fig:rq4_fidelity_resnet}
\end{figure}

As in \hyperref[sec:rq2]{RQ2} and \hyperref[sec:rq3]{RQ3}, we evaluate the $\lambda^\ast$ selected by highest \testrand accuracy in \hyperref[sec:rq1]{RQ1}.
Thus, RQ4 asks whether the robustness-selected models also yield more faithful explanations.
\autoref{fig:rq4_fidelity_resnet} summarizes the ResNet18 results for DDS with mean replacement under the robustness-selected \last.
We focus on mean-replacement DDS and ResNet18 in the main text and provide the corresponding \vit results, Gaussian perturbation results, and fidelity-oriented $\lambda$ selections on the supplementary website.
We summarize the main findings below.

\paragraph{Fidelity effects vary across architectures, methods, and shortcut settings.}
The two DDS perturbation variants yield similar qualitative conclusions, so we focus here on mean replacement.
Across settings, DDS changes depend strongly on architecture and explanation method.
For ResNet18, baseline DDS is often already high, especially for perturbation-based and surrogate explanations, leaving little room for improvement.
For \vit, lower baseline DDS in several settings leads to larger apparent gains.
Shortcut type and shortcut strength also do not determine the direction of the effect: patch, background, and color shortcuts each contain both improvements and decreases, and the correlation level \(p\) alone does not induce a monotonic DDS pattern.
Patch \(p=0.95\) is the clearest special case, where qualitative inspection suggests that most gradient explanations move away from the shortcut patch toward the foreground object, while the corresponding DDS changes remain mixed.
Thus, unlike robustness, reliance, or plausibility, fidelity is not governed by a clear global effect of \gls{egl}.

\paragraph{Attribution-constraining objectives improve fidelity selectively.}
For ResNet18, DDS improvements are concentrated mainly in gradient-based explanations under Invariance and Alignment.
This partially mirrors \hyperref[sec:rq3]{RQ3}, where the same objectives yield the clearest plausibility gains.
However, the fidelity effect is weaker and less systematic than the plausibility effect: Invariance and Alignment improve DDS in some settings, but do not define a stable ordering across datasets, architectures, and explanation methods.
For \vit, \gls{egl} more often improves DDS, particularly for initially weaker perturbation-based or surrogate explanations, but this again does not yield a consistent objective ranking.
Thus, objectives that directly optimize the explanation signal can improve perturbation-based fidelity in selected cases, but fidelity does not increase as reliably as plausibility.

\paragraph{Behavioral improvements do not imply higher fidelity.}
The clearest divergence appears for Sufficiency.
In \hyperref[sec:rq1]{RQ1} and \hyperref[sec:rq2]{RQ2}, Sufficiency produces some of the strongest robustness and foreground-reliance gains for separable shortcuts.
In RQ4, however, it often reduces DDS for ResNet18, especially in shortcut settings where it otherwise changes model behavior substantially.
For example, on Background \(p=0.95\), Sufficiency under the RQ1 selection improves \testrand accuracy and foreground preference, while DDS decreases for all six explanation methods.
This shows that models selected for robustness and foreground reliance do not necessarily yield higher perturbation-based fidelity.

\paragraph{Answer to RQ4.}
Overall, the experiments show that fidelity is not a direct consequence of either improved plausibility or improved shortcut robustness.
Invariance and Alignment can improve DDS for gradient-based explanations, but less consistently than they improve plausibility.
Conversely, Sufficiency can improve shortcut-shift robustness and foreground preference while reducing DDS.
This extends the VisFIS argument that plausibility and performance should be interpreted through the lens of faithfulness \cite{ying2022visfis}: in our controlled shortcut setting, fidelity forms a separate evaluation axis and cannot be inferred from robustness, reliance, plausibility, or rationale alignment alone.
At the same time, higher perturbation-based fidelity alone would not establish that the model relies on the intended evidence; it only indicates that the explanation better captures the model's current decision process.
The supplementary \vit results, Gaussian perturbation variant, and fidelity-oriented selections support the same qualitative conclusion.

%
%
%

\section{Discussion}
\label{sec:discussion}

Together, our results show that explanation-guided training can satisfy the visible rationale objective without reliably inducing the intended behavioral change.
Attribution-constraining objectives such as Invariance and Alignment most clearly improve explanation plausibility, especially for gradient-family explanations, but they provide weaker and less consistent gains in shortcut-shift robustness and foreground reliance.
This challenges a central motivation of \gls{egl}, namely that optimizing explanations should steer the model toward the intended evidence.
Viewed through Goodhart's law, rationale agreement becomes a proxy for the desired behavioral change: once optimized directly, it can improve as a measured explanation property without ensuring that the model's evidence use has changed accordingly.
This is not surprising in light of broader evidence that explanation maps can be manipulated without proportionally changing model behavior, for example through adversarial attacks or models trained to expose misleading explanation interfaces \cite{dombrowski2019explanations,dembinsky2026unifying,slack2021feature,zhang2020interpretable}.

The strongest behavioral correction comes from Sufficiency, which uses the rationale as data augmentation rather than as an explanation-alignment target.
By training on foreground-only inputs, Sufficiency changes the evidence available during optimization and can suppress shortcut reliance when shortcut and task evidence are separable.
This explains its strong effects for localized patches and many background shortcuts, where foreground masking removes shortcut evidence while preserving the object.
For global color shortcuts, the cue is distributed across the image and less cleanly separable from useful foreground information, leading to weaker and less stable improvements.
Thus, our results qualify VisFIS-style supervision \cite{ying2022visfis}: its objectives should not be interpreted as a uniformly additive package, since attribution constraints primarily improve plausibility whereas rationale-based data augmentation more directly changes model behavior.
At the same time, this should not be read as a general argument against \gls{egl}.
When rationales are causally grounded and well aligned with the data-generating process, even attribution-constraining objectives can be useful \cite{talies2026trustworthy}.
Our results instead caution against treating any plausible or human-annotated rationale as a sufficient ground truth for the model's intended decision process.

Fidelity further complicates this picture.
An illustrative case is ResNet18 on Patch $p=0.95$ under the robustness-selected \visa objective.
Here, several evaluation axes suggest a successful shift from shortcut-based to foreground-based evidence: \visa improves \testrand accuracy (\hyperref[sec:rq1]{RQ1}), increases foreground-only accuracy and reduces shortcut-only accuracy (\hyperref[sec:rq2]{RQ2}), and produces substantially more foreground-aligned explanations (\hyperref[sec:rq3]{RQ3}).
Yet DDS decreases across explanation methods (\hyperref[sec:rq4]{RQ4}).
This does not necessarily mean that explanation and behavior are completely misaligned.
Rather, \gls{egl} appears to move the explanation more strongly toward the foreground than it moves the model behavior itself.
The resulting explanation can become more plausible while becoming less aligned with the remaining sensitivities of the model, which is precisely what perturbation-based fidelity detects.
Conversely, baseline explanations can be highly faithful under perturbation because they identify the shortcut patch used by the shortcut-reliant model.
Perturbation fidelity should therefore be interpreted as a test of model sensitivity under a specified perturbation process, not as a generic certificate of semantic correctness.
This underlines a broader challenge in feature attribution evaluation, where different metrics can encode different assumptions about what constitutes a good explanation \cite{barr2023disagreement,dembinsky2026unifying}.

A practical implication is that \gls{egl} claims depend on the selected evaluation axis.
Our main analyses select \last by shortcut-shift robustness and then test whether other axes improve as well.
The supplementary model-selection analyses show the complementary view: selecting for reliance, plausibility, or fidelity can change which objective appears most successful, and improvements on one axis do not reliably recover gains on the others.\footnote{We encourage interested readers to inspect the supplementary material for a more holistic inspection: \supplink.}
Accordingly, \gls{egl} evaluations should state whether they target behavioral robustness, evidence reliance, explanation plausibility, fidelity, or a specific combination of these goals.
This caution extends beyond feature attribution and computer vision.
NLP rationales and concept-based models also show that constraining the visible explanation interface does not necessarily constrain the predictor to use only the intended evidence \cite{havasi2022addressing,lyu2024towards}.
The same issue is likely to become increasingly relevant for large language models, where natural-language rationales can be persuasive interfaces without necessarily exposing the causal basis of the prediction.

\subsubsection*{Limitations and future work.}
Our benchmark provides controlled access to shortcut structure, reference rationales, and diagnostic inputs, but necessarily simplifies real-world confounding.
Future work should test whether the observed trade-offs hold for natural shortcut datasets and domain-specific biases, especially in high-stakes settings such as medical imaging.
We focus on Saliency as the training-time guidance signal and on a fixed set of post-hoc explanation methods; other guidance signals, explanation methods, or attribution objectives may change the balance between robustness, reliance, plausibility, and fidelity.
Similarly, our combined \visa objective uses the individually selected coefficients for the four objective terms rather than a dedicated joint hyperparameter search, which may underestimate the attainable performance of the combined objective.
Future work could also make the proxy failure more explicit by training models toward intentionally uninformative or implausible rationales and testing how strongly explanation appearance can be decoupled from model behavior.
Finally, our results call for further work on whether perturbation-based fidelity metrics capture the right notion of faithfulness for explanation-guided models, and whether the community can develop more robust ways of assessing feature attribution fidelity.
This question is especially important as explanation-guided objectives and rationale-based supervision are extended from controlled vision benchmarks to broader settings, including language models and multimodal systems.

%
%
%
%
%

\section{Conclusion}

We presented a multi-axis evaluation of \gls{egl} under controlled shortcut bias, jointly analysing shortcut-shift robustness, diagnostic reliance, explanation plausibility, and explanation fidelity.
Our results show that direct explanation supervision can improve visible rationale agreement without reliably inducing the intended behavioral change.
Attribution-constraining objectives such as Invariance and Alignment primarily improve explanation plausibility, especially for gradient-family explanations, but provide weaker and less consistent robustness and reliance gains.
In contrast, rationale-based data augmentation, especially Sufficiency, more directly changes model behavior when shortcut evidence can be separated from task-relevant foreground information.

Fidelity further complicates this interpretation rather than resolving it.
Even when robustness, reliance, and plausibility jointly suggest a successful shift from shortcut-based to foreground-based evidence, perturbation-based fidelity can decrease.
Thus, \gls{egl} should not be judged from rationale agreement alone, but neither should fidelity be treated as a final certificate of semantic correctness.
Our results support a more cautious view: explanation-guided training can be useful, but its success depends on how the rationale enters training, whether it changes the evidence available to the model, and how behavioral and explanation-level evaluations are interpreted together.

\begin{credits}
\subsubsection{\ackname} 

\end{credits}

\bibliographystyle{splncs04}
\bibliography{bibliography}

\newpage
\appendix

\section{Dataset and Diagnostic Inputs}
\label{app:dataset}

\paragraph{Synthetic dataset construction.}
Each cFMNIST image is constructed by placing a Fashion-MNIST object on a CIFAR-10 background.
A binary foreground mask is obtained by thresholding the Fashion-MNIST image at pixel intensity $t=32$.
This mask defines the object support and the reference rationale $\tilde{E}$, while the pasted object retains the original Fashion-MNIST pixel intensities within the masked region.
Object and mask are transformed jointly by random resizing and rotation and then placed at a random image location.

\paragraph{Splits and shortcut correlations.}
Shortcut cues are implemented as a colored patch, a class-correlated CIFAR-10 background, or a global color-cast tint.
For each shortcut type and $p\in\{0.5,0.8,0.95\}$, the shortcut cue matches the foreground class with probability $p$.
Since cFMNIST has ten classes, $p=0.1$ corresponds to random shortcut assignment.
Each dataset contains $20{,}000$ training images, $5{,}000$ validation images, and $5{,}000$ images each for \testid and \testrand.
The \testid split uses the same shortcut correlation as training, whereas \testrand uses random shortcut assignment with $p=0.1$.
Unless stated otherwise, explanations in RQ3 and RQ4 are generated and evaluated on \testid; RQ1 and RQ2 use the evaluation splits and diagnostic inputs described in their respective sections.

\begin{figure}[t]
    \centering
    \includegraphics[width=1\linewidth]{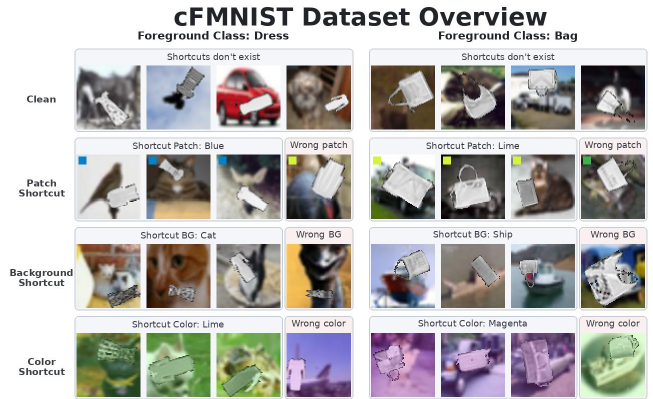}
    \caption{Examples from the cFMNIST dataset for the clean setting and the three shortcut variants: patch, background, and color-cast shortcuts.}
    \label{fig:cfmnist}
\end{figure}

\subsection{Foreground vs. Shortcut Diagnostics}
\label{app:data_fgbg}

To disentangle reliance on task-relevant and shortcut evidence, we evaluate trained models on foreground-only and shortcut-only diagnostic inputs.
The foreground-only input preserves the transformed Fashion-MNIST object and replaces all remaining pixels with the dataset mean background.

The shortcut-only input preserves only the shortcut signal of the corresponding regime.
For the clean setting, this yields an uninformative mean image.
For patch shortcuts, it contains the colored patch on the mean background.
For background shortcuts, it contains the CIFAR-10 background with the foreground region replaced by the dataset mean background, preventing object-shape leakage through a cut-out artifact.
For color shortcuts, it contains the global color-cast tint applied to the mean background.
Shortcut-only diagnostics use perfectly class-correlated shortcut cues ($p=1$), so their theoretical upper accuracy bound is $100\%$.

\begin{figure}[t]
    \centering
    \includegraphics[width=1\linewidth]{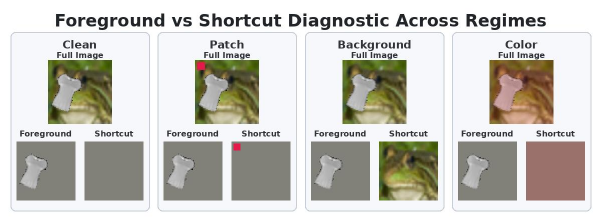}
    \caption{Foreground-only and shortcut-only diagnostic inputs used in \autoref{sec:rq2}. The foreground diagnostic preserves the Fashion-MNIST object on a mean background, whereas the shortcut diagnostic isolates the regime-specific shortcut cue.}
    \label{fig:cfmnist_regime}
\end{figure}

\section{Training and EGL Objectives}
\label{app:training_and_objectives}

\subsection{Model training}
\label{app:training_details}
\paragraph{Architectures and protocol.}
We use a ResNet18 \cite{he2016deep} trained from scratch and the ImageNet-pretrained \texttt{vit\_b\_16} implementation provided by PyTorch \cite{paszke2019pytorch}.
For each guidance objective, we perform a grid search over $\lambda\in\{0.05,0.1,0.25,0.5,1\}$.
Models are trained with five random seeds per nonzero $\lambda$.
The shared $\lambda=0$ baselines are trained with ten random seeds.

\paragraph{Training hyperparameters.}
Architecture-specific hyperparameters are fixed based on preliminary runs and held constant across all guidance regimes.
All models are trained for up to $80$ epochs with early stopping on the validation set.
For ResNet18, we use Adam with learning rate $10^{-3}$, weight decay $10^{-4}$, batch size $64$, and a cosine learning-rate schedule without warm-up; early stopping uses patience $12$.
For \vit, we use AdamW with learning rate $10^{-5}$, weight decay $0.05$, batch size $4$, label smoothing $0.05$, and a cosine learning-rate schedule after a $5$-epoch linear warm-up; early stopping uses patience $20$.
For guided runs, checkpoint selection is based on the full weighted validation objective, while the shared $\lambda=0$ baseline is selected by validation accuracy.

\subsection{Explanation guidance objectives}
\label{app:explanation_guidance}

\paragraph{Objective setup.}
Each individual \gls{egl} objective is added to the classification loss as
\[
    \mathcal{L}
    =
    \mathcal{L}_{\mathrm{CE}}(f(x),y)
    +
    \lambda \mathcal{L}_{\mathrm{EGL}} .
\]
For the combined \visa objective, we use the coefficients selected for the corresponding individual objectives:
\[
    \mathcal{L}_{\visa}
    =
    \mathcal{L}_{\mathrm{CE}}
    +
    \lambda_\mathrm{Suff}^{\ast}\mathcal{L}_\mathrm{Suff}
    +
    \lambda_\mathrm{Unc}^{\ast}\mathcal{L}_\mathrm{Unc}
    +
    \lambda_\mathrm{Inv}^{\ast}\mathcal{L}_\mathrm{Inv}
    +
    \lambda_\mathrm{Align}^{\ast}\mathcal{L}_\mathrm{Align}.
\]
During training, explanation-based objectives compute attributions with respect to the ground-truth class $y$.
For guided runs, checkpoint selection uses the full weighted validation objective.

\paragraph{Sufficiency.}
The sufficiency objective trains the model on a foreground-only view $x_e$, where the Fashion-MNIST object is preserved and non-foreground pixels are replaced by the mean background:
\[
    \mathcal{L}_\mathrm{Suff}
    =
    \mathrm{CrossEntropy}\big(f(x_e),y\big).
\]

\paragraph{Uncertainty.}
The uncertainty objective trains the model to be uninformative on a foreground-removed view $x_u$.
The foreground is covered by a circle enclosing its bounding box to avoid object-shape leakage through the removal boundary:
\[
    \mathcal{L}_\mathrm{Unc}
    =
    \mathrm{KLDivergence}\big(f(x_u),\mathrm{Uniform}(|\mathcal{Y}|)\big).
\]

\paragraph{Invariance.}
The invariance objective penalizes attribution mass outside the foreground rationale.
Let $\xi_y(f,x)$ denote the nonnegative, spatially normalized attribution map for the ground-truth class $y$.
Then
\[
    \mathcal{L}_\mathrm{Inv}
    =
    \left\lVert \xi_y(f,x)\odot(1-\tilde{E}) \right\rVert_1 .
\]

\paragraph{Alignment.}
The alignment objective directly matches the normalized attribution map to the normalized foreground mask:
\[
    \mathcal{L}_\mathrm{Align}
    =
    1-\mathrm{cos\text{-}sim}\big(\xi_y(f,x),\tilde{E}\big).
\]
For Invariance and Alignment, attributions are converted to nonnegative spatial mass by absolute channel-summing and normalization.

\paragraph{$\lambda$ selection.}
We tune $\lambda$ on the validation set using the same grid for all objectives.
To check whether the grid spans a plausible scale, we inspect loss magnitudes for Invariance.
Ross et al.~\cite{ross2017right} report that explanation regularization performs well when regularization and prediction losses are of comparable magnitude, with deviations of roughly one order still acceptable.
In our setting, $\mathcal{L}_\mathrm{Inv}$ is typically of the same order as the classification loss or up to two orders smaller, especially for \vit.
Empirically, the grid is sufficient: small nonzero $\lambda$ values already shift the optimized explanations, after which the effect plateaus.

\section{Post-hoc Explanations and Evaluation Metrics}
\label{app:evaluation_details}

\subsection{Explanation methods}
\label{app:explanation_methods}

For evaluation, explanations are computed for the predicted class $\hat{y}$, reflecting the deployment setting in which ground-truth labels are unavailable.
This convention is also natural for perturbation-based fidelity metrics, which evaluate the explanation of the model's own prediction.
Explanations and explanation-based evaluations are computed on \testid; checks on \testrand did not show meaningful qualitative differences.
Explanation hyperparameters were selected by manual inspection of representative examples rather than by optimizing any evaluation metric.
\textbf{Saliency}
\cite{simonyan2014deep} is computed as the absolute input gradient of the predicted-class logit.
\textbf{SmoothGrad}
\cite{smilkov2017smoothgrad} averages Saliency maps over $64$ noisy input copies with Gaussian noise standard deviation $0.15$ in normalized input space.
The averaged gradients are reduced by absolute channel-sum.
\textbf{Integrated Gradients}
\cite{sundararajan2017axiomatic} uses a zero baseline in normalized input space and $64$ integration steps with trapezoidal endpoint weighting.
Attributions are multiplied by the input-baseline difference and reduced by absolute channel-sum.
\textbf{Occlusion}
\cite{zeiler2014visualizing} replaces $12\times12$ patches with zero using stride $8$.
The attribution score is the change in the predicted-class logit; overlapping patch scores are averaged per pixel.
\textbf{LIME}
\cite{ribeiro2016should} uses SLIC superpixels with $32$ requested segments, compactness $10$, and $\sigma=0$.
We sample $256$ binary superpixel masks, replace masked regions with zero, evaluate the predicted-class logit, and fit a locally weighted ridge surrogate.
Superpixel coefficients are projected back to pixels.
\textbf{SHAP}
\cite{lundberg2017unified} is implemented as a KernelSHAP-style superpixel estimator using the same SLIC segmentation as LIME.
We sample $512$ coalition masks, include the all-on and all-off coalitions, apply the KernelSHAP weighting scheme, and fit a weighted linear surrogate to the predicted-class logit.
The resulting superpixel coefficients are projected back to pixels.

\subsection{Plausibility metrics}
\label{app:plausibility_metrics}

\paragraph{Importance Mass.}
Importance Mass measures the fraction of absolute attribution mass that lies inside the foreground mask:
\[
\mathrm{IM} = \frac{\sum_{i \in \tilde{E}} |\xi_i|}{\sum_i |\xi_i|}.
\]

\paragraph{AUROC.}
AUROC treats foreground pixels as positives and background pixels as negatives, and evaluates whether foreground pixels receive higher attribution scores than background pixels.

\paragraph{IoU AUC.}
For IoU AUC, each explanation map is min-max normalized per image to $[0,1]$. Binary explanation masks are obtained by thresholding this normalized map at $101$ thresholds between $0$ and $1$. We compute the IoU with the foreground mask at each threshold and report the normalized trapezoidal area under this IoU-threshold curve. Constant attribution maps are normalized to all zeros.

\subsection{Fidelity metrics}
\label{app:fidelity_metrics}

We evaluate explanation fidelity using Decaying Degradation Score (DDS) \cite{vsimic2022perturbation}.
It evaluates whether an explanation ranks features according to their relevance for the model output.
As for the explanations themselves, fidelity is computed with respect to the model's predicted class \(\hat{y}\) rather than the ground-truth label, so that the evaluation targets the decision actually explained by the attribution map.

DDS combines Most Relevant First (MoRF) and Least Relevant First (LeRF) perturbation curves.
In MoRF, features are iteratively perturbed from highest to lowest attribution score; in LeRF, they are perturbed from lowest to highest attribution score.
A faithful explanation should therefore cause a rapid loss of the original prediction under MoRF, while preserving it for longer under LeRF.
At each perturbation step \(i\), we measure predictive consistency with the original predicted class,
\[
    c_i = \mathds{1}\big[ f(x_i) = \hat{y} \big],
\]
rather than accuracy with respect to the ground-truth label.
Let \(M_i\) and \(L_i\) denote this consistency value after the \(i\)-th perturbation step for the MoRF and LeRF curves, respectively.
DDS is then computed as
\[
    \mathrm{DDS}
    =
    \sum_{i=1}^{n}
    (L_i - M_i)
    \left(\frac{n-i+1}{n}\right)^3 .
\]
The cubic decaying factor emphasizes early perturbation steps.
For pixel-based explanations, perturbation steps use fractions of \(0.05\) of all pixels; for LIME and SHAP, steps are taken over individual superpixels.

We evaluate two perturbation variants.
Mean replacement sets selected pixels to the fill value \(0\) in normalized input space.
Local Gaussian noise adds zero-mean Gaussian noise with \(\sigma=0.25\) to selected pixels.

\end{document}